\documentclass{article}
\usepackage{ijcai25}

\usepackage{times}
\usepackage{soul}
\usepackage{url}
\usepackage[hidelinks]{hyperref}
\usepackage[utf8]{inputenc}
\usepackage[small]{caption}
\usepackage{graphicx}
\usepackage{amsmath,amssymb,amsfonts}
\usepackage{amsthm}
\usepackage{booktabs}
\usepackage{algorithm}
\usepackage{algpseudocode}
\usepackage[switch]{lineno}

\usepackage{booktabs}
\usepackage{placeins}
\usepackage{tabularx}
\usepackage{makecell}
\usepackage{multirow} 
\usepackage{subcaption} 

\title{HGMP:Heterogeneous Graph Multi-Task Prompt Learning}

\author{
Pengfei Jiao$^1$
\and
Jialong Ni$^1$\and
Di Jin$^2$\and
Xuan Guo$^2$\And
Huan Liu$^1$\and
Hongjiang Chen$^1$\and
Yanxian Bi$^3$\thanks{Corresponding author.}\\
\affiliations
$^1$School of Cyberspace Security,Hangzhou Dianzi University\\
$^2$College of Intelligence and Computing,Tianjin University\\
$^3$CETC Academy of Electronics and Information Technology Group,China Academy of Electronic and Information Technology\\
\emails
\{nijialong, pjiao,huanliu,hchen\}@hdu.edu.cn,
\{jindi,guoxuan\}@tju.edu.cn,
biyanxian@cetc.com.cn
}

\begin{document}

\maketitle

\begin{abstract}
    The pre-training and fine-tuning methods have gained widespread attention in the field of heterogeneous graph neural networks due to their ability to leverage large amounts of unlabeled data during the pre-training phase, allowing the model to learn rich structural features. However, these methods face the issue of a mismatch between the pre-trained model and downstream tasks, leading to suboptimal performance in certain application scenarios. Prompt learning methods have emerged as a new direction in heterogeneous graph tasks, as they allow flexible adaptation of task representations to address target inconsistency. Building on this idea, this paper proposes a novel multi-task prompt framework for the heterogeneous graph domain, named HGMP. First, to bridge the gap between the pre-trained model and downstream tasks, we reformulate all downstream tasks into a unified graph-level task format. Next, we address the limitations of existing graph prompt learning methods, which struggle to integrate contrastive pre-training strategies in the heterogeneous graph domain. We design a graph-level contrastive pre-training strategy to better leverage heterogeneous information and enhance performance in multi-task scenarios. Finally, we introduce heterogeneous feature prompts, which enhance model performance by refining the representation of input graph features. Experimental results on public datasets show that our proposed method adapts well to various tasks and significantly outperforms baseline methods. 
\end{abstract}

\section{Introduction}

Heterogeneous graphs are powerful data structures that can simultaneously represent multiple types of nodes and edges, making them ideal for modeling complex relationships in various domains. For example, in social networks~\cite{cao2021knowledge}, heterogeneous graphs can model different types of entities such as users, posts, and comments, as well as various types of relationships, including friendships, comments, and likes. This flexibility enables heterogeneous graphs to capture rich and multifaceted interactions between entities, making them highly suitable for applications that require integration and analysis of diverse information, such as knowledge graphs~\cite{ji2021survey,hogan2021knowledge}, recommender systems~\cite{gao2023survey}, and citation networks.

Graph Neural Networks (GNNs)~\cite{yun2019graph,kipf2016semi,velivckovic2017graph} have demonstrated great performance in learning representations of graph-structured data by capturing local node relationships through message passing and aggregation. Heterogeneous Graph Neural Networks (HGNNs)~\cite{mao2023hinormer,fu2020magnn,yang2020heterogeneous} are an extension of GNNs, designed to handle the complexities of heterogeneous graphs. HGNNs incorporate both node and edge type information, allowing message passing to be more specialized and more effectively capture heterogeneous relationships. These models show improvements over standard GNNs in heterogeneous environments, particularly for tasks where understanding the relationships between entities depends heavily on node and edge types. However, the performance of both standard GNNs and HGNNs is limited by the scarcity of labeled data and the generalization ability on out-of-distribution  data~\cite{hu2019strategies,knyazev2019understanding,yehudai2021local,morris2019weisfeiler}. Furthermore, in multi-task scenarios, each task requires separate training, which further increases computational complexity.

To address these limitations, pre-training and fine-tuning strategies~\cite{dong2019unified} have been widely adopted. During pre-training, models learn general feature representations from large-scale unlabeled data, reducing the dependence on labeled datasets and improving generalization. Fine-tuning is then applied to adapt the model to specific tasks. However, the mismatch between pre-training objectives and downstream task requirements can negatively impact performance~\cite{liu2023graphprompt}, especially in scenarios with label imbalance or few-shot learning, where models struggle to generalize to new tasks or unseen data~\cite{zhang2022few}.

Prompt learning, a rapidly emerging technique in the machine learning community, provides an attractive alternative. It embeds task-specific prompts into the input data, guiding the model to perform specific tasks. For example, a sentiment analysis task like “IJCAI2025 will bring together top researchers from around the world. I feel so [MASK]” can be reformulated using a preset prompt (“I feel so [MASK]”). It is anticipated that a language model trained on masked word prediction might predict “[MASK]” as “enthusiastic” rather than “disappointed,” without requiring further parameter tuning for the new sentiment task, due to the model’s pre-training that embeds relevant knowledge to address this type of question. This demonstrates that prompt learning can effectively leverage existing knowledge, enabling the model to generalize across tasks with minimal additional training.

The success of prompt learning in NLP~\cite{li2021prefix,lester2021power} has inspired exploration into its application in graph learning. However, in graph learning, the use of prompt learning is still in its early stages. While some early works have explored the use of prompts in homogeneous graph settings~\cite{fang2024universal,sun2023all,sun2022gppt,liu2023graphprompt,tan2023virtual,huang2024prodigy,sun2023graph,zi2024prog}, these approaches do not address the unique challenges posed by heterogeneous graphs. Even recent prompt learning methods specifically designed for heterogeneous graphs still face some challenges~\cite{yu2024hgprompt,ma2024hetgpt}. First, aligning the pre-training objectives with the requirements of downstream tasks remains complex, as many early methods focus on link prediction tasks that connect data, neglecting the richness of heterogeneous data. Additionally, methods designed for single-task settings limit their applicability in multi-task scenarios. The main challenges in heterogeneous graph prompt learning can be summarized as: (C1) \textit{how to design unified graph templates for multi-task scenarios}, (C2) \textit{how to develop pre-training methods that adapt to heterogeneous data}, and (C3) \textit{how to design prompts that account for heterogeneity and feature diversity}.

To address these challenges, we propose the HGMP (Heterogeneous Graph Multi-task Prompt Learning) framework. Our approach rephrases node and edge-level tasks as graph-level tasks, aligning both downstream tasks and pre-training tasks in a common graph-level task space. This redefinition of tasks allows the model to undergo unified and effective training on graph-level objectives, capturing global graph structural features and enhancing the model's generalization ability. We introduce a heterogeneous pre-training strategy that combines subgraph extraction and augmentation for contrastive pre-training, fully leveraging the rich and diverse information in heterogeneous graphs. Furthermore, we design node-type-specific prompts using shared learnable vectors to modify node features, thereby enhancing the representation of input data and making it more suitable for the pre-training objectives.

In summary, the contributions of this paper are as follows:
\begin{itemize}  
    \item We propose a method that transforms node-level and edge-level tasks into graph-level tasks and introduces a flexible heterogeneous graph contrastive pre-training strategy to better capture the complexities of heterogeneous graph data.  
    \item We design the HGMP framework, specifically tailored for multi-task heterogeneous graph scenarios, which effectively bridges the gap between pre-training and downstream task objectives, enhancing the model's generalization ability.  
    \item We conduct comprehensive experiments on multiple public datasets, demonstrating the effectiveness, robustness, and adaptability of the HGMP framework across various task settings.  
\end{itemize}  

\section{Related Work}
\textbf{Heterogeneous Graph Neural Networks.} GNNs were initially designed for homogeneous graphs, focusing on node representation learning in graphs with consistent structures and attributes. To address the complexity of heterogeneous graphs, HGNNs emerged, enabling the modeling of diverse node and edge types. Representative works include HAN~\cite{wang2019heterogeneous}, which employs hierarchical attention mechanisms, HGT~\cite{hu2020heterogeneous}, a Transformer-based architecture for large-scale graphs, and Simple-HGN~\cite{lv2021we}, which enhances GAT~\cite{velivckovic2017graph} with learnable edge-type embeddings. Despite these advancements, HGNNs heavily rely on high-quality labeled data, which is often costly or infeasible to obtain, driving interest in pre-training techniques as alternatives to supervised learning.

\noindent\textbf{Graph Pre-training.} Graph pre-training aims to learn general knowledge from readily available data, reducing annotation costs. For homogeneous graphs, methods like GraphCL~\cite{you2020graph} utilize contrastive learning to exploit graph structures. In heterogeneous graphs, pre-training methods must handle diverse semantic relationships. Generative approaches reconstruct masked structures, while contrastive learning maximizes mutual information between positive sample pairs~\cite{jing2021hdmi,park2020unsupervised}, often outperforming generative methods. However, misalignment between pre-training objectives and downstream tasks can hinder the effective utilization of pre-trained knowledge.

\noindent\textbf{Prompt-based Learning On Graphs.} Prompt learning bridges the gap between pre-training and downstream tasks by introducing task-specific prompts, reducing fine-tuning demands and enhancing adaptability. Methods like GPPT~\cite{sun2022gppt}, GPF~\cite{fang2024universal}, and ProG~\cite{sun2023all} leverage prompts for task alignment and adaptation. In heterogeneous graphs, approaches such as HGPrompt~\cite{yu2024hgprompt} and HetGPT~\cite{ma2024hetgpt} use templates to support few-shot learning and align pre-training with downstream tasks. Nonetheless, current graph prompt learning methods often struggle with multi-task demands or rely on rigid templates, limiting flexibility and effectiveness.

\section{Preliminaries}

\textbf{Heterogeneous Graph.}
In graph data structures, a Heterogeneous Graph is an extended form designed to model complex networks with different types of nodes and edges. Compared to a homogeneous graph, a heterogeneous graph can more accurately represent diverse entities and their relationships in the real world. Formally, a heterogeneous graph can be defined as $G = (V, E, T_V, T_E)$, where $V$ represents the set of nodes, $E$ represents the set of edges, and $T_V$ and $T_E$ denote the sets of node types and edge types, respectively. In this definition, each node $v \in V$ is identified by a node type mapping $\phi(v) \in T_V$, and each edge $e \in E$ is associated with an edge type mapping $\psi(e) \in T_E$.

In addition to the structural aspects, nodes in a heterogeneous graph also have associated features. Different types of nodes often have different feature dimensions. Specifically, if a node $v$ belongs to type $t_1$, its feature vector has a dimension $d_{t_1}$, while if a node $u$ belongs to type $t_2$, its feature vector has a dimension $d_{t_2}$, where $d_{t_1}$ and $d_{t_2}$ are generally not equal.

\section{Method}
\subsection{Overview Framework}
\begin{figure*}[t]
\centerline{\includegraphics[width=0.9\textwidth]{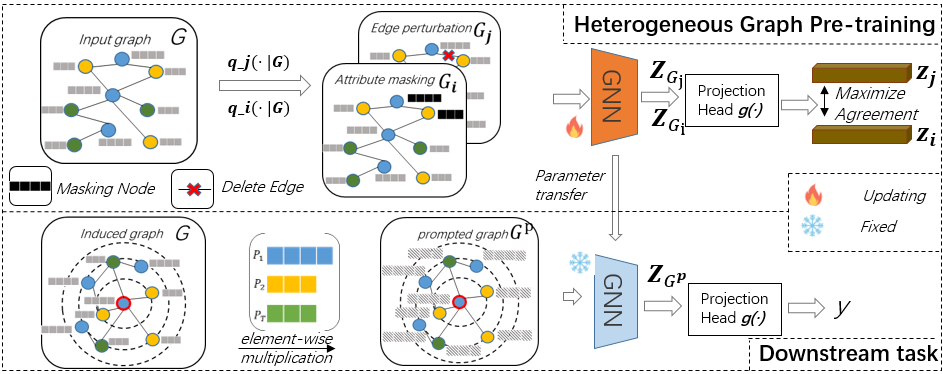}}
\caption{Overview of the HGMP architecture: The input graph is first subjected to heterogeneous graph augmentation, resulting in an augmented graph. Contrastive learning is then employed for pre-training, after which the GNN parameters are frozen. During downstream tasks, heterogeneous graph prompt features are multiplied with the node features of the input induced graph to perform graph augmentation, enabling alignment with the pre-training objective.}
\label{model}
\end{figure*}

In this paper, we aim to develop a graph-level pre-training task on a heterogeneous graph and learn a heterogeneous prompt vector capable of adjusting the original node features of the graph. By employing this heterogeneous prompt vector, we seek to bridge the gap between graph pre-training strategies and multiple downstream tasks, as well as to mitigate the challenges associated with transferring prior knowledge across different domains.To address this issue, we adopt a three-step approach. First, we unify all downstream tasks into a graph-level task format to minimize the gap between the pre-trained model and downstream task objectives. Next, recognizing that existing graph prompt learning methods struggle to effectively integrate contrastive pre-training strategies in the heterogeneous graph domain, we design a specialized graph-level contrastive pre-training strategy. This approach fully leverages heterogeneous information to enhance model performance in multi-task scenarios. Finally, we introduce heterogeneous feature prompts, which further improve model performance by optimizing the feature representation of input graphs. The overall framework of HGMF is shown in Fig~\ref{model}. 

\subsection{Reformulating Downstream Tasks}

\subsubsection{Challenges in Heterogeneous Graphs and Task Reformulation}
In homogeneous graphs, reformulating downstream tasks into a unified graph-level representation is a widely used approach to bridge the gap between pre-training and downstream tasks. By constructing $\tau$-ego networks as induced subgraphs and mapping node or edge labels to these subgraphs, node and edge classification problems can be reformulated as graph classification tasks. However, in heterogeneous graphs, this method faces new challenges due to the diverse types of nodes and edges, each carrying different semantics. Additionally, the semantic dependencies between these types are complex and cannot be captured by simple $\tau$-hop neighborhoods. The heterogeneity of these graphs makes it difficult to construct a consistent graph-level representation that can accommodate the full range of tasks.

\begin{figure}[htbp]  
    \centering  
    \begin{subfigure}[b]{0.3\textwidth}  
        \centering  
        \includegraphics[width=\textwidth]{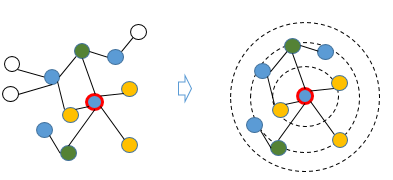} % 替换成你的图片文件名  
        \caption{Induced subgraphs for nodes}  
        \label{fig:image1}  
    \end{subfigure}  
    \hfill  
    \begin{subfigure}[b]{0.3\textwidth}  
        \centering  
        \includegraphics[width=\textwidth]{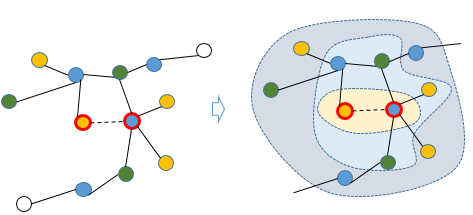} % 替换成你的图片文件名  
        \caption{Induced subgraphs for edge}  
        \label{fig:image2}  
    \end{subfigure}  
    \caption{Induced subgraphs for nodes and edges}  
    \label{fig:composite}  
\end{figure}

\subsubsection{Adapting Induced Subgraphs and Label Mapping}
To address these challenges, we adapt the task reformulation method to heterogeneous graphs by modifying the construction of induced subgraphs. For node-level tasks, the induced subgraph includes the target node, its $\tau$-hop neighbors, and all connecting edges along with their types, as show in Fig~\ref{fig:image1}. This subgraph captures both the structural context of the target node and the type-specific semantics of its neighboring nodes and edges. For edge-level tasks, the induced subgraph includes the two endpoints of the target edge and their respective $\tau$-hop neighbors, retaining all nodes, edges, and their types within this subgraph, as depicted in Fig~\ref{fig:image2}. This ensures that the semantic context surrounding the target edge is preserved, which is critical for accurate edge-level task performance. After constructing the induced subgraphs, we map the labels of the downstream tasks to these subgraphs. For node-level tasks, the label of the target node is assigned to its induced subgraph, transforming node classification into a graph classification task. Similarly, for edge-level tasks, the label of the target edge (e.g., relationship type) is assigned to its induced subgraph, turning edge classification into a graph classification problem. This mapping unifies node-level and edge-level tasks under a consistent graph-level framework, enabling knowledge transfer across tasks.

\subsubsection{Summary}
By reformulating both node-level and edge-level tasks into graph classification tasks, we provide a unified framework suitable for heterogeneous graphs. This method preserves the important type-specific semantics of heterogeneous graphs and lays a solid foundation for pre-training and downstream prompt-based learning in such graphs.

\subsection{Heterogeneous Graph Pre-training}

\subsubsection{Why Design Heterogeneous Graph Pre-training Tasks}

To unify pre-training and downstream tasks at the graph level, thereby maximizing the task subspace and fully exploring the potential relationships in heterogeneous graphs, we propose to adopt graph-level pre-training strategies during the pre-training phase. However, there are currently no pre-training methods specifically designed for graph-level tasks in heterogeneous graphs. This limitation prevents existing strategies from effectively adapting to graph-level downstream tasks, thereby restricting their performance and generalization. Moreover, contrastive learning has demonstrated superior performance in representation learning across various domains by constructing positive and negative sample pairs to enhance feature discrimination. Therefore, we propose a novel graph-level contrastive pre-training strategy for heterogeneous graphs to address these gaps and leverage the advantages of contrastive learning.
\subsubsection{How to Design Heterogeneous Graph Pre-training Tasks}

Heterogeneous graphs contain diverse node and edge types that play a critical role in preserving their structural integrity. For instance, in the ACM dataset, ``subject" nodes, despite comprising only 56 entities, connect to others through 3025 edges. To enhance model robustness without disrupting the graph structure, we propose two augmentation strategies: \textbf{Heterogeneous Node Masking Augmentation} and \textbf{Heterogeneous Edge Permutation Augmentation}, both guided by an augmentation ratio \( r \). This ratio determines how much of the node or edge set is altered, ensuring variability while maintaining structural integrity.

For \textbf{Heterogeneous Node Masking}, the adjusted augmentation ratio for each node type is computed as:
\[
\text{adjusted\_ratio}(i) = \frac{\text{count}(i)^2}{\sum_{j \in T_V} \text{count}(j)^2}
\]
where \( \text{count}(i) \) is the number of nodes of type \( i \), and \( T_V \) denotes all node types. Using this ratio, the final number of masked nodes is:
\[
\text{num\_to\_mask}(i) = r \cdot a(i) \cdot |V|
\]
where \( a(i) \) is the adjusted ratio for node type \( i \), \( r \) is the augmentation ratio, and \(|V|\) is the total number of nodes.

Similarly, for \textbf{Heterogeneous Edge Permutation}, the adjusted ratio for each edge type is:
\[
\text{adjusted\_ratio}(i) = \frac{\text{count}(i)^2}{\sum_{j \in T_E} \text{count}(j)^2}
\]
where \( \text{count}(i) \) is the number of edges of type \( i \), and \( T_E \) represents all edge types. The number of permuted edges is given by:
\[
\text{num\_to\_permute}(i) = r \cdot b(i) \cdot |E|
\]
where \( b(i) \) is the adjusted ratio for edge type \( i \), and \(|E|\) is the total number of edges.

After applying the augmentation strategies, the processing of augmented graphs involves three key steps. First, a GNN-based Encoder is used, incorporating a feature transformation layer to preprocess input features and extract graph-level representations \( \mathbf{Z}_{G_i}, \mathbf{Z}_{G_j} \). This flexible design ensures compatibility with various tasks. Second, a Projection Head maps high-dimensional representations into a smaller latent space using a two-layer MLP, which facilitates contrastive learning. Finally, the Contrastive Loss Function maximizes the similarity between positive pairs of augmented graphs while minimizing it for negative pairs. Given \( N \) graphs, \( 2N \) augmented graphs are generated per batch, and the loss is computed as:
\[
\mathcal{L}_{\text{con}} = - \sum_{i=1}^{N} \log \frac{\exp(\text{sim}(\mathbf{z}_i^+, \mathbf{z}_i^-)/\tau)}{\sum_{j=1}^{2N} \mathbb{1}_{[i \neq j]} \exp(\text{sim}(\mathbf{z}_i^+, \mathbf{z}_j)/\tau)}
\]
where \( \mathbf{z}_i^+ \) and \( \mathbf{z}_i^- \) are the positive and negative representations, and \( \tau \) is the temperature coefficient. By optimizing this loss, the model learns robust and discriminative graph-level representations that effectively support downstream tasks.

Through this integrated approach, our augmentation process efficiently combines the unique characteristics of heterogeneous graphs with contrastive pre-training, offering a unified and powerful graph representation learning framework.

\subsection{Heterogeneous Graph Prompt Feature (HGPF)}

\subsubsection{Why Design Heterogeneous Graph Prompt Feature (HGPF)}
In heterogeneous graphs, due to the diversity of node types, the feature dimensions of each node type are typically different. Even when the feature dimensions are consistent across nodes, the meanings of these dimensions often vary between different node types. For example, in a scholarly paper graph, author nodes might represent their research domains, while paper nodes contain publication information, making the meaning of the same feature dimension vary across different node types.

To effectively adjust the input graph for better adaptation to pre-trained models during downstream tasks, a mechanism must be designed to modify the feature representations of nodes according to their types. Therefore, we introduce the Heterogeneous Graph Prompt Feature (HGPF) to provide distinct feature prompts for each node type. This allows the adjustment of features of different node types, improving their alignment with pre-training tasks.

\subsubsection{How To Design Heterogeneous Graph Prompt Feature (HGPF)}

For each node type, we design a specific feature prompt associated with that type. Within each node type, the same feature prompt is shared. The dimensionality of the prompt matches the dimensionality of the corresponding node type's features. The role of this prompt is to adjust the node's feature representation by adding or multiplying it with the node's original feature.

Let \( \mathbf{x}_i \) denote the feature vector of node \( i \), where \( \mathbf{x}_i \in \mathbb{R}^{d_i} \) represents the feature vector of node \( i \), and \( T_i \) is the node type. The corresponding feature prompt is \( \mathbf{p}_{T_i} \in \mathbb{R}^{d_i} \). The adjustment process for the node feature is as follows:
% \[
% \hat{x}_i = x_i + p_{T_i}
% \]
% or
\[
\hat{\mathbf{x}}_i = \mathbf{x}_i \times \mathbf{p}_{T_i}
\]

Through this process, the node features are optimized by the adjustment based on their original features and the feature prompt, making them more suitable for the pre-training task.

\subsubsection{Why It Works?}

The effectiveness of HGPF is based on the prompt mechanism, which works by adjusting the node features of the input graph to better align with the pre-training task. In graph-based tasks, the flexibility of feature space operations is crucial to the effectiveness of prompting. By designing feature prompts, we make node features more task-relevant during pre-training, facilitating effective knowledge transfer.

Fang et al~\cite{fang2024universal}. demonstrated that for any graph \( G \) (with adjacency matrix \( A \) and node feature matrix \( X \)), an appropriate prompt token \( p^* \) can be learned such that the following equation holds:
\[
\varphi^*(A, X + p^*) = \varphi^*(g(A, X)) + O_{\varphi}
\]
Here, \( g(A, X) \) represents a transformation applied to the original graph (e.g., changing node features, adding/removing edges), and \( O_{\varphi} \) denotes the error between the graph augmented by the prompt and the graph after transformation. This error depends on the model's nonlinear layers and the quality of the learned prompt. With further optimization, the error can be significantly reduced.

In our approach, HGPF adjusts node features by customizing task-specific prompts for each node type. Unlike traditional methods, our feature prompts are combined with the original features of the nodes through multiplication. This multiplication method enables a more rapid adaptation to few-shot scenarios, particularly when dealing with attributed graphs, thereby enhancing the node representations for task alignment. This adaptive adjustment mechanism facilitates a more effective transfer of knowledge from the pre-trained model to the downstream task, significantly improving the performance of downstream tasks.
\begin{table}[ht]
    \centering
    \resizebox{\columnwidth}{!}{
        \begin{tabular}{lcccccc}
            \toprule
            Dataset   & \#Nodes &  \makecell{\#Node \\Types} & \#Edges  & \makecell{\#Edge \\Types} & Target & \#Classes \\
            \midrule
            ACM       & 10,942   & 4             & 547,872   & 8             & paper       & 3          \\
            IMDB      & 21,420   & 4             & 86,642   & 6             & movie      & 5          \\
            % Freebase  & 43,854  & 4             & 151,034 & 6            & movie        & 3          \\
            \bottomrule
        \end{tabular}
    }
        \caption{Summary of datasets.}
            \label{datasets}
\end{table}

\section{Experiments}
This section presents experimental results and analyses to demonstrate the superiority of our approach.
\subsection{Experimental Setup}
\textbf{Datasets.} We use two real-world HIN datasets, summarized in Table~\ref{datasets}. (1) \textit{ACM}~\cite{lv2021we} dataset is an academic network containing relationships among papers (P), authors (A), subjects (S), and terms (T), making it suitable for studying academic influence and knowledge discovery. (2) \textit{IMDB}~\cite{lv2021we} dataset serves as a benchmark for graph neural networks, involving movies (M), actors (A), directors (D), and keywords (K), with a wide range of relationships reflecting its suitability for tasks in the entertainment domain.

\noindent\textbf{Baselines.} We evaluate our method against baselines spanning five categories:
(1)Supervised GNNs: GCN~\cite{kipf2016semi}, GAT~\cite{velivckovic2017graph};
(2) GNNs with “pre-train, fine-tune”: GraphCL~\cite{you2020graph}, SimGRACE~\cite{xia2022simgrace};
(3) Supervised HGNNs: HGT~\cite{hu2020heterogeneous}, HAN~\cite{wang2019heterogeneous};
(4) HGNNs with “pre-train, fine-tune”: DMGI~\cite{park2020unsupervised}, HeCo~\cite{wang2021self}, HDMI~\cite{jing2021hdmi};
(5) HGNNs with “pre-train, prompt”: HGPROMPT~\cite{yu2024hgprompt}.

\noindent\textbf{Settings and parameters.} For the few-shot setting, we evaluate tasks under a 10-shot configuration across three categories: node classification, edge classification, and graph classification. The task construction mainly follows the approach of ProG~\cite{sun2023all} works. However, in heterogeneous graphs, target nodes are often connected to non-target nodes, whose learning value is relatively lower. This makes directly applying the ProG~\cite{sun2023all} approach, which combines both endpoints of edges as labels, less suitable in the heterogeneous domain.

To address this, we make specific adjustments: only the labels of nodes with endpoint types corresponding to target nodes are used as the induced graph's labels. The evaluation metrics include micro-F1 and macro-F1, ensuring a comprehensive assessment of the model's performance.For reproducibility, we provide the specific construction methods for each task and the detailed hyperparameter values in the supplement.

\subsection{Performance Evaluation on Multi-Tasks}

\subsubsection{Few-shot node classification}

% Please add the following required packages to your document preamble:
% \usepackage{booktabs}
% \usepackage{multirow}
% \usepackage[normalem]{ulem}
% \useunder{\uline}{\ul}{}
\begin{table}[]
\resizebox{\columnwidth}{!}{
\begin{tabular}{@{}l|ll|ll@{}}
\toprule
\multicolumn{1}{c|}{\multirow{2}{*}{Methods}} & \multicolumn{2}{c|}{ACM}                  & \multicolumn{2}{c}{IMDB}                  \\
\multicolumn{1}{c|}{}                         & Micro F1           & Macro F1           & Micro F1          & Macro F1           \\ \midrule
GAT                                           & 68.96±7.83          & 68.77±8.26          & 64.04±0.72          & 55.55±0.88          \\
GCN                                           & 81.50±1.62          & 81.46±1.70          & \ul{68.66±1.71}    & \ul{61.22±2.12}    \\ \midrule
GCL                                           & 66.73±1.89          & 66.61±1.94          & 65.08±0.92          & 56.81±1.15          \\
SimGRACE                                      & 55.00±2.79          & 54.71±2.86          & 64.63±0.36          & 56.27±0.45          \\ \midrule
HGT                                           & 61.78±7.81          & 57.12±11.19         & 64.64±1.30          & 56.26±1.61          \\
HAN                                           & 80.42±1.73          & 80.31±1.83          & 67.09±1.34          & 59.29±1.67          \\ \midrule
HeCo                                          & \ul{83.99±1.74}    & \ul{83.98±1.67}    & 66.18±1.08          & 58.16±1.32          \\
DMGI                                          & 65.42±2.56          & 64.07±2.81          & 59.41±1.60          & 49.80±1.97          \\ \midrule
HGPROMPT                                      & 77.29±1.81          & 75.40±2.80          & 61.76±0.78          & 52.71±0.95          \\
\textbf{HGMP}                                 & \textbf{84.60±3.18} & \textbf{84.54±3.19} & \textbf{68.76±0.82} & \textbf{61.34±1.00} \\ \bottomrule
\end{tabular}
}
\caption{Evaluation of node classification(\%). The best method is bolded and the runner-up is underlined.}
\label{node}
\end{table}

The classification results of the 10-shot nodes are shown in Table~\ref{node}. Based on the experimental results, we summarize our observations as follows:(1) HGMP outperforms all baseline methods across all datasets, demonstrating its effectiveness in node classification tasks under the few-shot setting within heterogeneous graph domains.(2) On the ACM dataset, HeCo, which also adopts a pre-training and fine-tuning strategy, achieves performance second only to HGMP and significantly surpasses homogeneous pre-training methods. This result highlights the necessity of fully leveraging heterogeneity information in pre-training methods for heterogeneous graphs. Furthermore, HeCo's performance, closely following that of HGMP, further validates the effectiveness of heterogeneous graph prompting mechanisms and the rationality of incorporating graph contrastive pre-training strategies that integrate heterogeneous information. Additionally, HAN, which utilizes meta-path strategies, also demonstrates competitive performance, showcasing its advantages in capturing structural and semantic information.(3) In few-shot scenarios, GCN significantly outperforms other homogeneous methods, indicating that models with smaller parameter sizes are more effective at avoiding overfitting when training data is limited, thereby achieving better generalization performance.

% Please add the following required packages to your document preamble:
% \usepackage{booktabs}

\subsubsection{Few-shot edge classification}

\begin{table}[]
\resizebox{\columnwidth}{!}{
\begin{tabular}{@{}l|ll|ll@{}}
\toprule
\multicolumn{1}{c|}{\multirow{2}{*}{Methods}} & \multicolumn{2}{c|}{ACM}                  & \multicolumn{2}{c}{IMDB}                  \\
\multicolumn{1}{c|}{}                         & Micro F1           & Macro F1           & Micro F1           & Macro F1           \\ \midrule
GAT                                           & 69.72±2.91          & 68.64±3.28          & 64.71±1.76          & 56.29±2.2           \\
GCN                                           & \ul{73.42±6.80}    & \ul{73.24±6.95}    & 66.36±1.05          & 58.32±1.30          \\ \midrule
GCL                                           & 61.82±3.07          & 61.34±3.60          & 64.68±1.05          & 56.56±1.28          \\
SimGRACE                                      & 50.85±2.05          & 50.19±2.30          & 63.23±1.23          & 54.45±1.53          \\ \midrule
HGT                                           & 66.61±9.85          & 61.21±13.68         & 63.49±0.75          & 54.77±0.92          \\
HAN                                           & 70.88±4.16          & 69.71±5.40          & 63.39±0.74          & 54.65±0.92          \\ \midrule
HeCo                                          & 70.13±1.45          & 70.30±1.43          & \ul{66.50±1.31}    & \ul{58.80±1.62}    \\
DMGI                                          & 57.09±2.08          & 55.47±3.50          & 63.05±1.58          & 54.55±1.95          \\ \midrule
HGPROMPT                                      & 61.74±5.77          & 58.33±7.35          & 60.43±2.46          & 50.98±3.02          \\
\textbf{HGMP}                                 & \textbf{77.33±2.53} & \textbf{77.44±2.53} & \textbf{67.65±0.83} & \textbf{60.19±1.01} \\ \bottomrule
\end{tabular}
}
\caption{Evaluation of edge classification(\%). The best method is bolded and the runner-up is underlined.}
\label{edge}
\end{table}

The results of the 10-shot edge classification task are shown in the Table~\ref{edge}. Based on the experimental results, we summarize the following observations:
(1) HGMP outperforms all baseline methods in edge-type tasks, demonstrating its broad adaptability to few-shot settings in tasks within heterogeneous graph domains.
(2) Since HeCo and DMGI only model the features of the target nodes, they fail to capture the complete information required for the edge classification task, thus losing their previous second-best performance in node classification tasks. GCN, once again, benefits from its smaller parameter size, achieving success in the few-shot setting due to its ability to avoid overfitting.
\subsubsection{Few-shot graph classification}

% Please add the following required packages to your document preamble:
% \usepackage{booktabs}
% \usepackage{multirow}
% \usepackage[normalem]{ulem}
% \useunder{\uline}{\ul}{}
\begin{table}[]
\resizebox{\columnwidth}{!}{
\begin{tabular}{@{}l|ll|ll@{}}
\toprule
\multicolumn{1}{c|}{\multirow{2}{*}{Methods}} & \multicolumn{2}{c|}{ACM}                  & \multicolumn{2}{c}{IMDB}                  \\
\multicolumn{1}{c|}{}                         & Micro F1           & Macro F1           & Micro F1           & Macro F1           \\ \midrule
GAT                                           & 98.07±0.67          & 98.06±0.67          & \ul{51.39±4.34}    & \ul{50.8±5.02}     \\
GCN                                           & \ul{98.16±0.58}    & \ul{98.16±0.58}    & 49.42±3.32          & 47.38±3.28          \\ \midrule
GCL                                           & 86.92±2.09          & 87.00±2.00          & 42.21±2.22          & 40.50±2.53          \\
SimGRACE                                      & 70.44±3.05          & 70.31±2.93          & 34.78±2.15          & 33.36±2.73          \\ \midrule
HGT                                           & 89.18±8.76          & 88.21±11.19         & 33.99±6.69          & 25.85±8.22          \\
HAN                                           & 96.92±0.96          & 96.91±0.97          & 27.97±4.96          & 22.77±8.71          \\ \midrule
HeCo                                          & 97.55±0.42          & 97.55±0.42          & 46.36±1.78          & 45.30±1.76          \\
DMGI                                          & 89.02±2.68          & 88.96±2.67          & 41.25±3.68          & 40.53±3.57          \\ \midrule
HGPROMPT                                      & 91.13±1.54          & 90.94±1.72          & 32.32±1.93          & 31.01±1.76          \\
\textbf{HGMP}                                 & \textbf{98.30±0.45} & \textbf{98.30±0.45} & \textbf{53.73±3.38} & \textbf{52.88±3.60} \\ \bottomrule
\end{tabular}
}
\caption{Evaluation of graph classification(\%). The best method is bolded and the runner-up is underlined.}
\label{graph}
\end{table}

\begin{table*}[]
\centering
\begin{tabular}{@{}l|cc|c|c|c@{}}
\toprule
Methods   & \begin{tabular}[c]{@{}c@{}}Heterogeneous \\      graph augmentation\end{tabular} & \begin{tabular}[c]{@{}c@{}}Heterogeneous \\      graph feature\end{tabular} & \begin{tabular}[c]{@{}c@{}}Node classification\\      ACM    IMDB\end{tabular} & \begin{tabular}[c]{@{}c@{}}Edge classification\\      ACM   IMDB\end{tabular} & \begin{tabular}[c]{@{}c@{}}Graph classification\\      ACM   IMDB\end{tabular} \\ \midrule
VARIANT 1 & ×                                                                                & ×                                                                           & 66.73   65.08                                                                  & 61.82   64.68                                                                 & 86.92    42.21                                                                 \\
VARIANT 2 & ×                                                                                & \checkmark                                                                           & 81.71   67.94                                                                  & 74.93   65.86                                                                 & 97.80    52.46                                                                 \\
VARIANT 3 & \checkmark                                                                                & ×                                                                           & 68.83   65.38                                                                  & 66.04   65.33                                                                 & 89.89    44.86                                                                 \\
HGMP      & \checkmark                                                                                & \checkmark                                                                           & \textbf{84.60   68.76}                                                         & \textbf{77.33   67.65}                                                        & \textbf{98.30    53.73}                                                        \\ \bottomrule
\end{tabular}
\caption{Variants used in ablation study, and corresponding results in Micro F1 (\%) on node,edge and graph classification.}
\label{ablation}
\end{table*}

The results of the 10-shot graph classification task are shown in Table~\ref{graph}. Based on the experimental results, we summarize the following observations:
(1) On the ACM dataset, all baseline methods improved significantly in graph-type tasks, due to the high concentration of same-label target nodes during subgraph construction. HGMP outperformed all baselines, demonstrating its superiority in multi-task few-shot settings within heterogeneous graphs.
(2) The IMDB dataset, being a multi-label multi-class classification task, performed well in node and edge classification. However, in the graph classification task, to adhere to the previous task setup, we could only select one label, converting the problem into a single-label multi-class classification task. This conversion resulted in significant information loss, leading to a noticeable drop in performance across all baselines on the IMDB dataset in the graph classification task.
(3) The strong performance of all baselines in the graph classification task highlights that the knowledge obtained from pretraining on graph-level tasks is better suited for downstream graph classification tasks. This further underscores the necessity of unifying task designs at the graph level.
(4) GAT demonstrated a performance recovery, indicating that as the graph size increases slightly, attention-based models can regain competitive performance. However, HeCo and DMGI, as they only model the features of target nodes, remain unable to capture complete graph-level information and consequently fell out of the top three in this task.
\begin{table}[]
\centering
\resizebox{\columnwidth}{!}{
\begin{tabular}{@{}l|l|c|c|c@{}}
\toprule
\multirow{2}{*}{Backbone} & \multirow{2}{*}{method} & \multirow{2}{*}{\begin{tabular}[c]{@{}c@{}}Node   Classification\\      Micro F1 Macro F1\end{tabular}} & \multirow{2}{*}{\begin{tabular}[c]{@{}c@{}}Edge   Classification\\      Micro F1 Macro F1\end{tabular}} & \multirow{2}{*}{\begin{tabular}[c]{@{}c@{}}Graph   Classification\\      Micro F1 Macro F1\end{tabular}} \\
                          &                         &                                                                                                               &                                                                                                               &                                                                                                                \\ \midrule
\multirow{2}{*}{GCN}      & SUPERVISED              & 81.50       81.46                                                                                             & 73.42       73.24                                                                                             & 98.16       98.16                                                                                              \\
                          & HGMP                    & \textbf{84.60       84.54}                                                                                    & \textbf{77.33       77.44}                                                                                    & \textbf{98.30       98.30}                                                                                     \\ \midrule
\multirow{2}{*}{GAT}      & SUPERVISED              & 68.96       68.77                                                                                             & 69.72       68.64                                                                                             & \textbf{98.07       98.06}                                                                                     \\
                          & HGMP                    & \textbf{73.18       73.37}                                                                                    & \textbf{70.18       70.20}                                                                                    & 95.78       95.78                                                                                              \\ \midrule
\multirow{2}{*}{HGT}      & SUPERVISED              & 61.78       57.12                                                                                             & 66.61       61.21                                                                                             & 89.18       88.21                                                                                              \\
                          & HGMP                    & \textbf{68.25       67.62}                                                                                    & \textbf{72.47       70.23}                                                                                    & \textbf{89.49       88.35}                                                                                     \\ \bottomrule
\end{tabular}
}
\caption{Evaluation of different backbones for NC,EC and GC tasks(\%) on ACM. SUPERVISED means end-to-end training.}
\label{backbone}
\end{table}

\subsection{Flexibility on backbones}
To validate the adaptability and performance of the proposed HGMP model across different backbones, we conducted experiments using GCN, GAT, and HGT on node classification, edge classification, and graph classification tasks. HGMP consistently outperformed traditional supervised methods in node and edge classification. For GCN, HGMP achieved improvements of approximately 3\% and 4\% in node and edge classification, respectively, while maintaining comparable performance in graph classification. With GAT, HGMP excelled in edge classification, outperforming the baseline by nearly 1\%, though it showed a slight performance decrease in graph classification. For HGT, HGMP provided noticeable gains in node and edge classification, albeit with modest improvements in graph classification. These results confirm HGMP's adaptability and effectiveness in few-shot heterogeneous graph tasks, leveraging graph augmentation and prompt features to enhance performance.

\begin{figure}[t]
\centerline{\includegraphics[width=\linewidth]{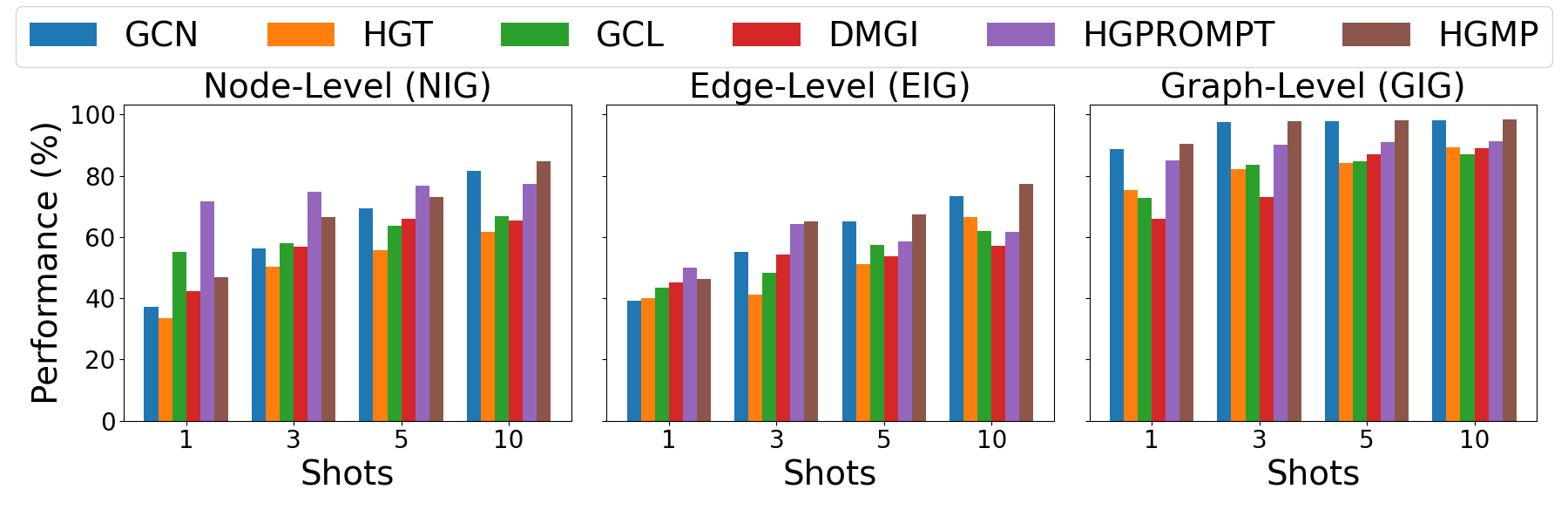}}
\caption{Impact of shots on ACM.}
\label{shots}
\end{figure}

\subsection{Performance with different shots}
To analyze the impact of shot numbers, we conducted experiments on the ACM dataset for node classification (NC), edge classification (EC), and graph classification (GC). The results, shown in Figs~\ref{shots}, indicate that HGMP generally outperforms baselines, especially in EC and GC tasks. For NC, HGPROMPT excels in few-shot scenarios (fewer than 5 shots) due to its optimized class prototype but loses its edge as sample size increases, with HGMP showing stronger robustness and generalization. In EC and GC tasks, HGMP consistently achieves superior performance, attributed to its graph-level training objectives and stable adaptability across settings.

\subsection{Ablation study}
To evaluate the contribution of each component in HGMP, we conducted an ablation study comparing the full model with its variants, as shown in Table~\ref{ablation}. All variants were trained using the reconstruction task template based on induced subgraphs to ensure fair comparisons. The results demonstrate that each component significantly enhances performance.

Comparing Variant 1 with Variant 3 highlights that the absence of heterogeneous graph augmentation causes performance degradation, validating the importance of optimizing augmentation strategies to capture complex structural information. Similarly, the comparison between Variant 1 and Variant 2 shows that heterogeneous prompt features provide substantial performance gains, effectively constructing task-relevant representations and improving classification accuracy.

Overall, HGMP outperforms all variants, demonstrating the synergistic benefits of combining heterogeneous graph augmentation and prompt features. This design aligns induced subgraphs in downstream tasks with pretraining objectives, enhancing generalization and overall performance.

\section{Conclusions}
This paper proposes HGMP, a novel multi-task prompt framework that bridges pre-trained models and downstream tasks in heterogeneous graphs by unifying downstream tasks into a graph-level format and incorporating a graph-level contrastive pre-training strategy leveraging heterogeneous information. Experimental results demonstrate its significant advantages over existing methods, particularly in multi-task scenarios. While effective for attributed graphs, HGMP performs suboptimally on unattributed graphs, highlighting a need for improved adaptability. Future work will focus on addressing this limitation, offering new perspectives for heterogeneous graph learning and advancing multi-task learning and practical graph applications.

\section*{Acknowledgements}
This work was supported in part by the Zhejiang Provincial Natural Science Foundation of China under Grant LDT23F01015F01, in part by the Key Technology Research and Development Program of the Zhejiang Province under Grant No. 2025C1023 and in part by the National Natural Science Foundation of China under Grants 62372146.

%% The file named.bst is a bibliography style file for BibTeX 0.99c
\bibliographystyle{named}
\bibliography{ijcai25}

\end{document}